\documentclass[pdflatex,sn-mathphys-num]{sn-jnl}% Math and Physical Sciences Numbered Reference Style
\usepackage{graphicx}%
\usepackage{multirow}%
\usepackage{amsmath,amssymb,amsfonts}%
\usepackage{amsthm}%
\usepackage{mathrsfs}%
\usepackage[title]{appendix}%
\usepackage{xcolor}%
\usepackage{textcomp}%
\usepackage{manyfoot}%
\usepackage{booktabs}%
\usepackage{algorithm}%
\usepackage{algorithmicx}%
\usepackage{algpseudocode}%
\usepackage{listings}%
\usepackage{longtable}
\usepackage{adjustbox}
\usepackage{graphicx}
\usepackage{subcaption} % provides \subcaptionbox

\theoremstyle{thmstyleone}%
\theoremstyle{thmstyletwo}%

\theoremstyle{thmstylethree}%

\begin{document}

% \title[Article Title]{Center-based Entropy Classifier}
\title[Article Title]{CEDL: Centre-Enhanced Discriminative Learning for Anomaly Detection}
%%=============================================================%%
%% GivenName	-> \fnm{Joergen W.}
%% Particle	-> \spfx{van der} -> surname prefix
%% FamilyName	-> \sur{Ploeg}
%% Suffix	-> \sfx{IV}
%% \author*[1,2]{\fnm{Joergen W.} \spfx{van der} \sur{Ploeg} 
%%  \sfx{IV}}\email{iauthor@gmail.com}
%%=============================================================%%

\author*[1]{\fnm{Zahra} \sur{Zamanzade Darban}}\email{zahra.zamanzadeh@monash.edu}

\author*[2]{\fnm{Qizhou} \sur{Wang}}\email{mike.wang@unimelb.edu.au}
% \equalcont{These authors contributed equally to this work.}

\author[3]{\fnm{Charu C.} \sur{Aggarwal}}\email{charu@us.ibm.com}
% \equalcont{These authors contributed equally to this work.}

\author[1]{\fnm{Geoffrey~I.} \sur{Webb}}\email{geoff.webb@monash.edu}
% \equalcont{These authors contributed equally to this work.}

\author[1]{\fnm{Ehsan} \sur{Abbasnejad}}\email{ehsan.abbasnejad@monash.edu}
% \equalcont{These authors contributed equally to this work.}

\author[1]{\fnm{Mahsa} \sur{Salehi}}\email{mahsa.salehi@monash.edu}
% \equalcont{These authors contributed equally to this work.}

\affil*[1]{\orgdiv{DSAI}, \orgname{Moansh University}, \orgaddress{ \city{Melbourne}, \state{Victoria}, \country{Australia}}}

\affil[2]{\orgdiv{School of CIS}, \orgname{The University of Melbourne}, \orgaddress{\city{Parkville}, \state{Victoria}, \country{Australia}}}

\affil[3]{\orgdiv{IBM T. J. Watson Research Center}, \orgname{IBM}, \orgaddress{\city{Yorktown Heights}, \state{NY}, \country{USA}}}

%%==================================%%
%% Sample for unstructured abstract %%
%%==================================%%

\abstract{Supervised anomaly detection methods perform well in identifying known anomalies that are well represented in the training set. However, they often struggle to generalise beyond the training distribution due to decision boundaries that lack a clear definition of normality. Existing approaches typically address this by regularising the representation space during training, leading to separate optimisation in latent and label spaces. The learned normality is therefore not directly utilised at inference, and their anomaly scores often fall within arbitrary ranges that require explicit mapping or calibration for probabilistic interpretation. To achieve unified learning of geometric normality and label discrimination, we propose Centre-Enhanced Discriminative Learning (CEDL), a novel supervised anomaly detection framework that embeds geometric normality directly into the discriminative objective. CEDL reparameterises the conventional sigmoid-derived prediction logit through a centre-based radial distance function, unifying geometric and discriminative learning in a single end-to-end formulation. This design enables interpretable, geometry-aware anomaly scoring without post-hoc thresholding or reference calibration. Extensive experiments on tabular, time-series, and image data demonstrate that CEDL achieves competitive and balanced performance across diverse real-world anomaly detection tasks, validating its effectiveness and broad applicability. The source code is available on~\url{https://github.com/zamanzadeh/CEC}.
}


%%\pacs[JEL Classification]{D8, H51}

%%\pacs[MSC Classification]{35A01, 65L10, 65L12, 65L20, 65L70}

\maketitle

\section{Introduction}
Anomaly detection \cite{grubbs1969procedures, chandola2009anomaly, pang2021deep} is the task of identifying data samples that differ significantly from the majority. It remains a popular and active research problem due to its high relevance across a wide range of application domains, such as cyber-physical systems, finance, healthcare, and industrial monitoring. Anomaly detection methods are traditionally designed as unsupervised learning problems on unlabeled data \cite{breunig2000lof, ramaswamy2000efficient, scholkopf2001estimating, tax2004support, leung2005unsupervised, liu2008isolation}. They assume that the unlabeled dataset is largely composed of normal samples with few anomalies, identifying anomalies as points that deviate significantly from the majority distribution. However, as these methods operate in the original feature space or rely on fixed, non-learnable transformations, their effectiveness is limited when handling complex, high-dimensional natural data.

% Detecting anomalies, observations that deviate from the regular behavior of a system, remains a core problem across domains such as cyber-physical systems, finance, healthcare, and industrial monitoring. Modern deployments must operate under three persistent constraints: (i) anomalies are rare and heterogeneous, often spanning multiple modes; (ii) labeled anomalies are scarce or unavailable; and (iii) models must remain robust when normal patterns drift over time. While deep neural networks (DNNs) have advanced representation learning for complex time series, the decision layer that converts representations into anomaly judgments is typically inherited from standard classification practice (e.g., linear or affine logits trained with binary cross-entropy). This mismatch can yield brittle boundaries when the underlying class geometry is radial or clustered around one or a few centers, as is common in one-class and novelty-detection settings.

To effectively learn discriminative features for anomaly detection on complex, high-dimensional inputs, recent studies have adopted deep learning approaches that use diverse neural network backbones \cite{lecun2015deep} for feature extraction, followed by dedicated anomaly scoring modules. Many of these methods \cite{xu2015learning, andrews2016detecting, erfani2016high, ruff2018deep, xu2023deep} still rely on unsupervised training to avoid the need for labelled data. However, unsupervised learning methods are often insufficient to characterise the anomalies of interest or to provide a clear definition of normality, leading to high false positive rates or unsatisfactory sensitivity to anomalies. In contrast, supervised methods (i.e, these utilise labelled training data) generally \cite{pang2018learning, devnet, dou2020enhancing,  pang2021explainable, ding2021few} achieve substantially stronger detection performance on anomalies that resembling the training instances. This approach is feasible in many well-established domains where labelled normal samples and labelled anomalies are available. However, these methods often struggle to generalise to anomaly types that are not well represented by the training data \cite{prenet, wang2025openset}. This limitation arises because discriminative binary classification losses, such as binary cross-entropy (BCE), fit the decision boundary based on known anomalies but do not enforce a compact boundary around the normal class, making the model unable to capture precise normality and prone to errors in anomaly identification under train–test distribution shifts.

Several recent approaches \cite{sad, prenet, wang2025openset} have been proposed to alleviate the limited generalisation of supervised training in anomaly detection by enforing stronger normality that constrain and regularise the discriminatively learned decision boundaries. For instance, Deep SAD \cite{sad} leverages labelled anomaly samples to calibrate hypersphere-based one-class objectives, while PReNet \cite{prenet} exploits pairwise relationships to improve generalisation to anomalies that are not represented in the training data. Nevertheless, these methods primarily focus on constraining the latent space or refining existing one-class formulations through auxiliary supervision. Moreover, their anomaly scores often lie within an arbitrary range that does not correspond to a meaningful probabilistic value, making it difficult to define consistent thresholds for decision-making in real-world applications where detection sensitivity must be controlled. No existing work has achieved end-to-end optimisation that integrates a discriminative objective with latent geometric structure and can also achieve direct anomaly scoring without post-hoc thresholding. 

To address this research gap, we propose the Center-Enhanced Discriminative Learning (CEDL), a supervised anomaly detection framework that embeds geometric normality directly into the discriminative formulation, resulting in a single, coherent learning formulation capable of probabilistic anomaly scoring without requiring explicit thresholding. Our key insight is that geometry-defined normality can be expressed as a radial logit, providing a principled alternative to the conventional log-linear decision boundary, which is less suitable for anomaly detection due to its hyperplane assumption over known training anomalies. 

Specifically, CEDL reparameterises the sigmoid-derived class probability with a radial logit computed from the distance to the centroid of normal instances. Within a weighted cross-entropy formulation, this maps hyperspherical distances to probabilistic outputs, allowing geometry-based and interpretable anomaly scores. This enables simultaneous optimisation of representation geometry and label discrimination through the final output layer. Intuitively, CEDL aligns the classifier with the geometry of normality, where instances close to the learned centre are deemed normal, while those further away incur increasing anomaly evidence through a monotone distance-to-probability mapping. CEDL is backbone and modality agnostic, fully differentiable, and can be seamlessly integrated with diverse backbone architectures for anomaly detection across different data modalities, to perform end-to-end training and anomaly scoring. Through extensive experiments on tabular, time series, and image data, CEDL-enabled detection models achieve superior anomaly detection performance, validating the feasibility of embedding one-class geometric constraints into the label discrimination objective, while demonstrating its high effectiveness and broad applicability.

\begin{itemize}
    \item We propose Centre-Enhanced Discriminative Learning (CEDL), a novel supervised anomaly detection framework that embeds geometric normality directly into the discriminative learning objective via a radial logit centred on the distribution of normal samples, unifying discriminative and geometric learning in a single end-to-end formulation.
    \item We show that a centre-based radial distance function can reparameterise the conventional sigmoid-derived prediction logit, enabling geometry-aware normality learning within a label-supervised discriminative framework. At inference time, the learned monotonic distance-to-probability mapping offers interpretable, geometry-aware anomaly scoring that requires no reference data calibration or manual thresholding.
    \item Extensive experiments on tabular, time-series, and image anomaly detection benchmarks demonstrate that CEDL delivers competitive and balanced performance across diverse real-world datasets, validating its wide applicability and the feasibility of embedding geometric normality modelling within a classification objective.
\end{itemize}

% This paper makes the following contributions:
% \begin{itemize}
%     \item A geometry-aligned framework for deep supervised anomaly detection. We introduce CEC, a radial-logit decision layer that replaces linear logits with a distance-based mapping to probability, better matching the geometry of one-class and novelty detection.
%     \item Backbone-controlled evaluation. We perform head-to-head comparisons where CEC, PreNet, DeepSAD, DevNet, DeepSVDD, and BCE share the same deep backbone and identical training hyperparameters, isolating the impact of the decision function.
%     \item Broad empirical validation. Across UTS/MTS and tabular benchmarks, we evaluate using best F1, AU-ROC, and AU-PR, demonstrating consistent gains where anomaly structure is heterogeneous or multi-modal.
% \end{itemize}

\section{Related Works}
\label{sec:rw}

\subsection{Classis Anomaly Detection Methods}
Early anomaly detection methods \cite{chandola2009anomaly, pang2021deep} are mostly unsupervised and rely on statistical or geometric principles rather than complex, learnable transformations. They include density-based models such as Gaussian Mixture Models \cite{bishop2006pattern} and Kernel Density Estimation \cite{davis2011remarks}, distance-based and clustering-based methods like k-Nearest Neighbour \cite{ramaswamy2000efficient} and HiCS \cite{keller2012hics}, hypersphere-based one-class approaches such as One-Class SVM \cite{scholkopf2001estimating} and SVDD \cite{tax2004support} that enclose normal data within compact boundaries, and isolation-based methods like Isolation Forest\cite{liu2008isolation} that separate anomalies through recursive random partitioning. These methods are valued for their simplicity and robustness but assume anomalies are rare and identifiable through geometric or statistical deviations. However, their reliance on fixed feature representations limits their ability to capture complex, high-dimensional inputs, making them unable to learn anomaly-discriminative features and often leading to inaccurate definitions of the normal class.

\subsection{Deep Anomaly Detection}
To alleviate these limitations, numerous deep learning-based methods \cite{pang2021deep} have been explored to leverage the strong representation learning capacity of deep neural networks for anomaly detection. Among them, autoencoder-based \cite{hinton2006reducing, chen2017outlier, zong2018deep, erfani2016high, gong2019memorizing, zhang2019deep} approaches reconstruct input data through a bottleneck architecture, assuming that anomalies, which are poorly captured in the learned latent space, yield higher reconstruction errors. Generative adversarial network-based methods \cite{goodfellow2014generative, zenati2018adversarially, schlegl2019f, zhang2022deep}  utilise adversarial learning to capture the distribution of normal data, where anomalies are detected as samples that the generator fails to reproduce or the discriminator recognises as inconsistent with the learned normal patterns. One-class based methods \cite{ruff2018deep, goyal2020drocc, chen2022deep, sohn2020learning}, extend classical one-class formulations by deep embeddings that compactly enclose normal data in a latent space. Although these approaches leverage deep architectures to capture complex and high-dimensional data, they remain unsupervised and rely on implicit anomaly patterns that are insufficient to define true normality and actual anomalies.

Supervised deep anomaly detection methods use labelled data for direct supervision, making them more effective at detecting anomalies similar to those used for training. These approaches \cite{pang2018learning, ding2021few, zhang2019deep, sad, prenet} utilise a limited number of labelled anomalies and have shown to substantially improve detection performance. Positive and unlabeled (PU) learning methods \cite{luo2018pu, sansone2018efficient} leverage a limited number of labelled anomalies together with unlabeled data that mostly contain normal samples. Many of these methods assume similar anomaly distributions between training and testing, resulting in decision boundaries that fit known anomalies but have limited generalisation to others. Recent supervised methods \cite{prenet, ding2022catching, wang2025openset} are proposed to improve generalisation to novel anomaly types at inference by introducing additional objectives alongside label discrimination. Our work follows this line of research but is distinct in that it embeds geometric normality directly into the discriminative objective, achieving unified optimisation of representation geometry and label-based discriminative training without explicitly regularising the representation space.

\section{Proposed method}
\label{sec:model}
% \textbf{Problem definition: }Given a dataset of samples with representations $r \in \mathbb{R}^D$ and binary labels $y \in \{0,1\}$, the problem is to learn a classifier that separates normal samples ($y=0$) from anomalous samples ($y=1$).  
\textbf{Problem definition: } We consider a supervised anomaly detection setting with a labelled training dataset $D_{\text{train}} = {(x_i, y_i)}{i=1}^{N}$ with $N$ samples, where $x_i \in \mathbb{R}^d$ denotes an input sample and $y_i \in {0, 1}$ indicates whether it is normal ($y_i = 0$) or anomalous ($y_i = 1$). An end-to-end anomaly scoring network is defined as $f_{\theta}(x_i) = \eta(\psi(x_i))$, where the representation learner $\psi(\cdot): \mathbb{R}^d \rightarrow \mathbb{R}^m$ maps inputs into an $m$-dimensional representation space, and the scoring module $\eta(\cdot): \mathbb{R}^m \rightarrow [0,1]$ produces anomaly scores. 
The objective is to learn $f_{\theta}$ such that anomalous samples receive higher scores than normal ones, allowing a threshold to be drawn for identifying anomalies.

% The Center--Entropy Classifier (CEC) achieves this by replacing the linear logit with a radial logit based on the Euclidean distance of $r$ to a normal center $c$, and optimizes a weighted binary cross-entropy loss with class weights $w_0,w_1>0$ to account for class imbalance.

\subsection{Center-Enhanced Deep Supervised Anomaly Detection}
We aim to combine the strengths of discriminative learning and geometry-based compact representation learning within a unified objective to achieve robust anomaly detection through discriminative training with labels without explicitly regularising the representation space. To this end, we propose Center-Enhanced Discriminative Learning (CEDL), a supervised framework that embeds centre-based one-class representation learning into an entropy-based binary classification objective. The key idea is to reparameterise the sigmoid-derived class probability used in standard entropy loss with a radial logit computed from the distance to the centroid of normal instances. By minimising the CEDL's loss on the training data, the separation between normal and anomalous classes is enforced, and the compactness of the normal class region is simultaneously established within the representation space, thereby enhancing discriminability between normal and anomalous instances and yielding a more precise definition of normality. For inference, the centre of the learned normal representations serves as a reference for distance-based anomaly scoring without relying on a hyperplane decision boundary. Given $N$ labelled examples $\{(x_i, y_i)\}_{i=1}^{N}$, the objective can be defined as
\begin{equation}
\mathcal{L}_{\text{CEDL}} =
-\frac{1}{N}
\sum_{i=1}^{N}
\Big[
y_i \log(s_i)
+
(1 - y_i)\log(1 - s_i)
\Big], 
\end{equation}
where $s_i = f(\|\phi(x_i) - c\|)$ is derived from the distance between a sample’s representation and the centre of normal samples $c$ in the latent space. We name this CEDL Loss. The objective resembles the binary cross-entropy formulation but is conceptually different, as it operates on feature–centre distances instead of predicted probabilities derived from sigmoid, embedding geometric structure into the learning process. For normal samples, the objective encourages their feature representations to move closer to the centre, forming a compact cluster that defines the region of normality. In contrast, anomalous samples are pushed away from the centre, increasing their radial distance and creating clear separation from the normal cluster. This geometry ensures that  normal and anomalous samples are moved in opposite directions, leading to a stable and interpretable representation space where normality is tightly defined and anomalies deviates from the normal area.

\subsection{Property of Center-Enhanced Discriminative Loss}
The CEDL loss comes with several desirable properties that contribute to anomaly detection compared to conventional discriminative objective such as BCE. Let $r_i = \phi(x_i)$ denote the latent representation of sample $x_i$ and $c$ the learnable centre representing the normal class. The CEDL loss defines a radial logit as
\begin{equation}
    a_i = \frac{\alpha}{\sqrt{D}} \|r_i - c\|_2,
\end{equation}
and applies a weighted BCE-with-logits objective:
\begin{equation}
\mathcal{L}_{\text{CEDL}} =
\frac{1}{N}\sum{i=1}^{N}
\big[
w_1 y_i \, \text{softplus}(-a_i)
	•	w_0 (1 - y_i) \, \text{softplus}(a_i)
\big],
\label{{eq:CEDL-softplus}}
\end{equation}
where $w_0, w_1 > 0$ are class weights and $\alpha > 0$ controls the scaling of the radial distance. The gradient of the loss with respect to $r$ can be calculated as:
\begin{align*}
\nabla_r \mathcal{L}_{\mathrm{CEDL}}
& = w_y\;(\sigma(a)-y)\;\frac{\partial a}{\partial r}
\label{eq:CEDL-grad} \\
& = w_y\;(\sigma(a)-y)\;\frac{\alpha}{\sqrt{D}}\;\frac{r-c}{\|r-c\|_2},
\end{align*}
where \(\sigma(\cdot)\) is the sigmoid function. Based on this gradient, we observe the following property.

\bigskip

\noindent\textbf{Radial gradient direction}. This gradient always lies in the radial direction $(r_i - c)/\|r_i - c\|$, ensuring that each update depends only on the sample’s position relative to the centre. Unlike a linear BCE classifier, which moves samples along a single projection vector, CEDL adjusts them directly along the radial axis.

\bigskip

\noindent \textbf{Monotonic updates for normal and anomalous samples.} The gradient update for the representation $r_i$ is \(\Delta r = -\eta\,\nabla_r \mathcal{L}\), where \(\eta>0\) is the learning rate. Taking the dot product of this update with the radial vector $(r_i - c)$ gives:
\begin{equation}
 \Delta r_i \cdot (r_i - c) =
	•	\eta w_{y_i} (\sigma(a_i) - y_i)
\frac{\alpha}{\sqrt{D}}
\|r_i - c\|_2.
\end{equation}
This relation determines how the distance between each sample and the centre changes:
For normal samples $(y_i = 0)$, since $\sigma(a_i) \in (0,1)$, we have $(\sigma(a_i) - 0) > 0$, so $\Delta r_i \cdot (r_i - c) < 0$. The distance $\|r_i - c\|$ decreases, pulling normals toward the centre. Similarly, for anomalous samples, $(y_i = 1)$, since $(\sigma(a_i) - 1) < 0$, we get $\Delta r_i \cdot (r_i - c) > 0$. The distance increases, pushing anomalies outward. Hence, every gradient step monotonically adjusts the distance in the correct direction, stable radial separation between normal and anomalous samples during training.

\begin{algorithm}[t]
\footnotesize
\caption{\texttt{CEDL Training}($\mathcal{D}$, $\alpha$, $w_0$, $w_1$, $\eta$, $E$)}
\label{alg:cedl}
\begin{algorithmic}[1]
  \renewcommand{\algorithmicrequire}{\textbf{Input:}}
  \renewcommand{\algorithmicensure}{\textbf{Output:}}
    \Require Training set $\mathcal{D} = \{(x_i, y_i)\}_{i=1}^N$, scale $\alpha$, class weights $w_0, w_1$, learning rate $\eta$, epochs $E$
    \Ensure Representation network $\phi^R$
    
    \State Initialize $\phi^R$ parameters, optimizer with learning rate $\eta$
    \State Set center $c \gets \mathbf{0}\in\mathbb{R}^D$\hfill 
    \State $best\_loss \gets \infty$
    
    \For{$epoch = 1$ to $E$}
        \State $total\_loss \gets 0$
        \For{each mini-batch $(X,Y)$ from $\mathcal{D}$}
            \State $R \gets \phi^R(X)$ \hfill
            \State $a(r) \gets \tfrac{\alpha}{\sqrt{D}}\|r-c\|_2$ for each $r\in R$
            \State Compute $\ell_{\text{CEDL}}(r,y)$ using Eq.~(\ref{{eq:CEDL-softplus}})
            \State $\mathcal{L}_{\text{CEDL}} \gets \tfrac{1}{|R|}\sum \ell_{\text{CEDL}}(r,y)$
            
            \State Backpropagate $\nabla \mathcal{L}_{\text{CEDL}}$ and update $\phi^R$ parameters
            \State $total\_loss \gets total\_loss + \mathcal{L}_{\text{CEDL}}$
        \EndFor
        
        \State $avg\_loss \gets total\_loss / |\mathcal{D}|$
        \If{$avg\_loss < best\_loss$}
            \State $best\_loss \gets avg\_loss$
            \State Save current $\phi^R$ state
        \EndIf
    \EndFor
    
    \State \textbf{Return} $\phi^R$
\end{algorithmic}
\end{algorithm}

\subsection{Anomaly Detection using CEDL}
\textbf{Training.} Algorithm \ref{alg:cedl} shows the pseudo code for CEDL loss over the labelled dataset $\mathcal{D} = \{(x_i, y_i)\}{i=1}^N$. At each iteration, a mini-batch of samples is passed through the network to obtain latent representations $r_i = \phi(x_i)$. For each representation, the radial logit $a(r_i) = \tfrac{\alpha}{\sqrt{D}}\|r_i - c\|2$ is computed with respect to the centre c, which may be fixed or learnable. The per-sample loss $\ell_{\text{CEDL}}(r_i, y_i)$ is evaluated using the weighted BCE-with-logits formulation, and the batch loss $\mathcal{L}_{\text{CEDL}}$ is obtained by averaging across the mini-batch. Gradients of $\mathcal{L}_{\text{CEDL}}$ are then backpropagated to update the parameters of $\phi^R$ using the learning rate $\eta$. The process repeats for $E$ epochs, and the model state with the lowest average loss is retained as the final representation network. This procedure directly encourages compact clustering of normal representations around the centre while pushing anomalous representations outward, leading to stable convergence and geometrically interpretable embeddings.

\bigskip

\noindent \textbf{Inference.} During inference, the trained representation network $\phi^R$ encodes each input $x$ into a latent representation $r = \phi^R(x)$. The anomaly score is computed as the Euclidean distance between the representation and the learned centre $c$ as $a(r) = \|r - c\|_2$. A larger distance indicates greater deviation from the normal region and thus a higher likelihood of anomaly. This provides a simple yet interpretable inference mechanism that directly reflects the geometry of the learned representation space. Unlike purely discriminative classifiers that rely on fixed decision boundaries in logit space, which may not preserve feature-space relationships or handle heterogeneous anomaly patterns, this approach measures deviation from the normal centre, enabling continuous, geometry-aware ranking. 

\section{Experiment}
\label{sec:experiment}

\subsection{Datasets} \label{sec:datasets}
We evaluate the proposed method on a diverse collection of tabular and time-series datasets to examine its generality across data modalities. 

\bigskip

\noindent \textbf{Tabular datasets.} Following the benchmark  established in previous works \cite{devnet, prenet}, we adopt twelve tabular datasets widely used in deep anomaly detection research: Thyroid, Campaign, Donors, Fraud, Backdoor, Census, Fuzziness (FUZ), Recursion (REC), DoS, W7a, News20, and CelebA.
These datasets span multiple domains, including medical diagnosis, cybersecurity, finance, and image attributes, and exhibit diverse feature dimensionalities and anomaly ratios.

\bigskip

\noindent \textbf{Time-series datasets.} To evaluate the performance of CEDL on time series anomaly detection, we adopt four benchmark datasets encompassing both univariate and multivariate settings: Yahoo \cite{yahoods}, Key Performance Indicators (KPI), Mars Science Laboratory (MSL), and Soil Moisture Active Passive (SMAP). MSL and SMAP are multivariate spacecraft telemetry datasets collected from NASA missions, containing labelled anomalies derived from incident reports for spacecraft monitoring systems. Yahoo comprises 367 hourly time series with anomaly labels, and we focus on the A1 benchmark, which includes 67 univariate series representing real production traffic data from Yahoo services. KPI is sourced from real-world Internet service scenarios and consists of time series capturing various operational metrics such as response time, page views, CPU utilisation, and memory usage. For all datasets, we follow the original training and testing splits.

\bigskip

\noindent \textbf{Image datasets.} The image anomaly detection experiments are conducted on three widely used benchmarks: MNIST \cite{mnist}, Fashion-MNIST \cite{fmnist}, and CIFAR-10 \cite{cifar}. These datasets contain visually diverse samples spanning handwritten digits, clothing items, and natural objects, respectively.

\subsection{Evaluation Metrics}
We evaluate model performance using three standard metrics: Area Under the Receiver Operating Characteristic (AUROC), Area Under the Precision–Recall Curve (AUPR), and Best F1-score. AUROC and AUPR are holistic measures of the model’s ability to distinguish between normal and anomalous samples across all possible thresholds. AUROC evaluates the trade-off between true and false positive rates, making it suitable for assessing the model’s overall ranking quality. AUPR measures precision–recall performance under class imbalance, emphasising detection reliability when anomalies are rare. Best F1-score complements these threshold-independent metrics by evaluating performance at the optimal decision threshold, providing a practical measure of detection accuracy in real-world settings.

\subsection{Baselines} \label{sec:baselines}
We compare the proposed CEDL method against five representative baseline methods for deep anomaly detection: Deep SVDD \cite{ruff2018deep}, Deep SAD \cite{sad}, DevNet \cite{devnet}, PReNet \cite{prenet}, and BCE. Deep SVDD learns a compact hypersphere enclosing normal samples in latent space, forming the basis for many supervised extensions. Deep SAD incorporates labelled anomalies to jointly attract normal samples to the centre and repel anomalies outward. DevNet employs a deviation-based loss that measures how far each prediction deviates from the normal distribution, offering a probabilistic view of abnormality. PReNet learns pairwise relational features and anomaly scores by predicting relationships between randomly sampled instance pairs, enhancing discrimination and improving generalisation to both seen and unseen anomalies. Finally, the BCE baseline applies the standard binary cross-entropy loss with sigmoid outputs as a fully supervised classifier for anomaly separation.

\subsection{Experimental Setup} \label{sec:setup}
To evaluate generalisation under train–test distribution shifts, we adopt a systematic rotation protocol. Class 0 is designated as the normal class, while each of the remaining classes is alternately treated as the known anomaly during training. For example, in the MNIST dataset with ten classes, we conduct nine experiments where classes 1–9 each serve as the known anomaly class in turn. In each experiment, models are trained on labelled samples from class 0 (normal) and one known anomaly class, while the remaining classes appear only during testing. This setup creates a controlled distribution shift, requiring models to detect both known anomalies and generalise to unseen anomaly types not observed during training to achieve robust overall performance.

\subsection{Implementation Details}
\noindent \textbf{Tabular experiments.}
For tabular experiments, all models use a fully connected MLP encoder with ReLU activations and a bounded \texttt{tanh} output. Concretely, the encoder comprises three hidden layers of sizes $1000$, $256$, and $64$, followed by a linear projection to a $32$-dimensional representation and a final \texttt{tanh} to keep embeddings in a compact range. Inputs are the raw feature vectors. When using SVMLight files, sparse matrices are densified prior to batching. Datasets provided in normalised/binarised form are used as supplied, with no additional feature engineering. Data are split using a stratified $60/40$ train/test split with a fixed seed ($42$), preserving the original class imbalance in the training set. Training uses the Adam optimiser with a learning rate of $0.0001$, batch size $64$, and $100$ epochs. We used a static centre at the origin; class imbalance is handled via a weight ratio computed from the training split, $w = N_{\text{normal}}/N_{\text{anomalous}}$.

\noindent \textbf{Time series experiments.}
For time series experiments, we use a 1D convolutional sequence encoder (adapted from prior residual Conv1D designs) that maps fixed-length windows to compact embeddings consumed by the CEDL objective. Each input sequence is segmented into windows of length $100$ with stride $1$; channels are standardised per window using the training split's statistics. We used a static centre at the origin. We perform a chronological split to avoid leakage: the first half of each series is used for training and the second half for testing. Training uses Adam with a learning rate of $0.0001$, batch size $32$, and $200$ epochs. The CEDL loss is applied with a static centre and imbalance weighting $w = N_{\text{normal}}/N_{\text{anomalous}}$ computed from the training windows. Reproducibility is ensured by fixing the random seed to $42$ and enabling deterministic CuDNN settings.

\noindent \textbf{Image experiments.} For image experiments, all models adopt a LeNet-style CNN encoder with MaxPool, BatchNorm, and Leaky ReLU activations. The architecture is adjusted according to dataset complexity, where MNIST and Fashion-MNIST use two convolutional layers followed by one fully connected layer with 32-dimensional representations, while CIFAR-10 employs three convolutional layers and one fully connected layer with 128-dimensional representations. Training is performed using the Adam optimiser with a learning rate of 0.0001 and a batch size of 64. MNIST and Fashion-MNIST models are trained for 50 epochs, whereas CIFAR-10 is trained for 100 epochs. The training data consist of $10\%$ normal samples and $2.5\%$ known anomaly samples. For evaluation, $10\%$ anomalies are randomly sampled from both the known and unseen anomaly classes to maintain a balanced test distribution and prevent overrepresentation of anomalies.

\subsection{Anomaly Detection Performance} \label{sec:compare}
{\small
\begin{longtable}{lcccccc}
\caption{Comparison of CEDL with the baselines on twelve tabular datasets. Bold values indicate the best performance for each metric.}
\label{tab:tabular_results_long} \\
\toprule
\textbf{Method} & \textbf{PReNet} & \textbf{DeepSAD} & \textbf{DevNet} & \textbf{DeepSVDD} & \textbf{BCE} & \textbf{CEDL} \\
\midrule
\endfirsthead
\multicolumn{7}{c}{\tablename\ \thetable\ -- \textit{Continued from previous page}} \\
\toprule
\textbf{Method} & \textbf{PReNet} & \textbf{DeepSAD} & \textbf{DevNet} & \textbf{DeepSVDD} & \textbf{BCE} & \textbf{CEDL} \\
\midrule
\endhead
\midrule
\multicolumn{7}{r}{\textit{Continued on next page}} \\
\endfoot
\bottomrule
\endlastfoot
\multicolumn{7}{c}{\textbf{Thyroid}} \\
\cmidrule(lr){1-7}
AUROC & 0.782 & \textbf{0.997} & 0.641 & 0.558 & 0.983 & 0.994 \\
AUPR & 0.276 & \textbf{0.963} & 0.142 & 0.087 & 0.938 & 0.953 \\
F1 & 0.321 & 0.907 & 0.159 & 0.079 & \textbf{0.916} & 0.912 \\
\midrule
\multicolumn{7}{c}{\textbf{Campaign}} \\
\cmidrule(lr){1-7}
AUROC & 0.827 & 0.884 & 0.470 & 0.386 & \textbf{0.906} & 0.852 \\
AUPR & 0.459 & 0.511 & 0.130 & 0.104 & 0.513 & \textbf{0.576} \\
F1 & 0.511 & 0.521 & 0.150 & 0.113 & 0.569 & \textbf{0.576} \\
\midrule
\multicolumn{7}{c}{\textbf{Donors}} \\
\cmidrule(lr){1-7}
AUROC & \textbf{1.000} & \textbf{1.000} & \textbf{1.000} & 0.121 & \textbf{1.000} & \textbf{1.000} \\
AUPR & 0.997 & \textbf{1.000} & \textbf{1.000} & 0.033 & \textbf{1.000} & \textbf{1.000} \\
F1 & 0.991 & \textbf{1.000} & \textbf{1.000} & 0.000 & \textbf{1.000} & \textbf{1.000} \\
\midrule
\multicolumn{7}{c}{\textbf{Fraud}} \\
\cmidrule(lr){1-7}
AUROC & \textbf{0.981} & 0.963 & 0.849 & 0.842 & 0.937 & 0.971 \\
AUPR & 0.690 & 0.764 & 0.128 & 0.483 & 0.499 & \textbf{0.779} \\
F1 & 0.796 & 0.792 & 0.234 & 0.629 & 0.472 & \textbf{0.816} \\
\midrule
\multicolumn{7}{c}{\textbf{Backdoor}} \\
\cmidrule(lr){1-7}
AUROC & 0.972 & 0.992 & 0.960 & 0.654 & 0.918 & \textbf{0.995} \\
AUPR & 0.876 & \textbf{0.984} & 0.852 & 0.180 & 0.840 & 0.981 \\
F1 & 0.905 & 0.945 & 0.836 & 0.305 & 0.911 & \textbf{0.952} \\
\midrule
\multicolumn{7}{c}{\textbf{Census}} \\
\cmidrule(lr){1-7}
AUROC & 0.854 & 0.887 & \textbf{0.909} & 0.428 & 0.764 & 0.906 \\
AUPR & 0.348 & 0.455 & 0.449 & 0.053 & 0.410 & \textbf{0.514} \\
F1 & 0.408 & 0.496 & 0.477 & 0.048 & \textbf{0.555} & 0.529 \\
\midrule
\multicolumn{7}{c}{\textbf{FUZ}} \\
\cmidrule(lr){1-7}
AUROC & 0.883 & \textbf{0.937} & 0.880 & 0.552 & 0.560 & \textbf{0.937} \\
AUPR & 0.175 & 0.487 & 0.186 & 0.101 & 0.148 & \textbf{0.489} \\
F1 & 0.211 & 0.449 & 0.175 & 0.094 & 0.214 & \textbf{0.467} \\
\midrule
\multicolumn{7}{c}{\textbf{REC}} \\
\cmidrule(lr){1-7}
AUROC & 0.964 & \textbf{0.998} & 0.962 & 0.619 & 0.990 & 0.995 \\
AUPR & 0.757 & \textbf{0.988} & 0.762 & 0.237 & 0.971 & 0.984 \\
F1 & 0.748 & 0.952 & 0.671 & 0.249 & \textbf{0.954} & 0.963 \\
\midrule
\multicolumn{7}{c}{\textbf{DoS}} \\
\cmidrule(lr){1-7}
AUROC & 0.946 & \textbf{0.999} & 0.970 & 0.503 & 0.991 & 0.993 \\
AUPR & 0.895 & \textbf{0.994} & 0.905 & 0.306 & 0.979 & 0.986 \\
F1 & 0.863 & \textbf{0.963} & 0.777 & 0.281 & 0.958 & 0.962 \\
\midrule
\multicolumn{7}{c}{\textbf{W7a}} \\
\cmidrule(lr){1-7}
AUROC & 0.874 & 0.931 & 0.866 & 0.596 & 0.847 & \textbf{0.955} \\
AUPR & 0.490 & 0.756 & 0.648 & 0.040 & 0.685 & \textbf{0.794} \\
F1 & 0.559 & 0.741 & 0.599 & 0.051 & \textbf{0.785} & 0.794 \\
\midrule
\multicolumn{7}{c}{\textbf{News20}} \\
\cmidrule(lr){1-7}
AUROC & 0.956 & 0.909 & 0.960 & 0.294 &0.958& \textbf{0.990} \\
AUPR & 0.625 & 0.329 & 0.632 & 0.108 &0.887 & \textbf{0.953} \\
F1 & 0.721 & 0.426 & 0.734 & 0.182 & 0.892 & \textbf{0.900} \\
\midrule
\multicolumn{7}{c}{\textbf{CelebA}} \\
\cmidrule(lr){1-7}
AUROC & \textbf{0.958} & 0.956 & 0.938 & 0.621 & 0.500 & 0.937 \\
AUPR & 0.303 & 0.361 & 0.239 & 0.065 & 0.364 & \textbf{0.405} \\
F1 & 0.399 & 0.401 & 0.276 & 0.109 & \textbf{0.403} & 0.443 \\
\midrule
\multicolumn{7}{c}{\textbf{Rank}} \\
\cmidrule(lr){1-7}
AUROC & 4.167 & 1.917 & 3.833 & 5.833 & 3.182 & \textbf{1.333} \\
AUPR & 3.000 & 1.917 & 3.583 & 5.833 & 3.727 & \textbf{1.833} \\
F1 & 4.000 & 2.583 & 4.333 & 5.750 & 2.273 & \textbf{1.250} \\
\end{longtable}}
\bigskip
%%=============================================%%
%% For presentation purpose, we have included  %%
%% \bigskip command. Please ignore this.       %%
%%=============================================%%

\subsubsection{Anomaly Detection on Tabular Data}
The preliminary results in Table \ref{tab:tabular_results_long} demonstrates that the proposed CEDL method consistently achieves superior performance across all tabular datasets, with the highest AUROC, AUPR, and F1 scores in nearly all cases. CEDL attains the best overall rank across all three metrics. Given that the baseline performance on tabular data is already high, these improvements highlight CEDL’s capability to detect challenging anomalous instances that other models fail to capture. This demonstrates strong discriminative accuracy and robustness across data from diverse domains. The performance gain over the classic BCE baseline is substantial, validating that incorporating centre-distance–based radial logic into cross-entropy training produces a synergistic effect that enhances representation learning and anomaly discrimination. This design leads to a geometry-aware anomaly scoring mechanism that generalises more effectively under train–test distribution shifts.

\subsubsection{Anomaly Detection on Time Series Data}
\begin{table}[t]
\centering
\caption{Performance comparison on time series anomaly detection datasets.
Bold values indicate the best result for each dataset and metric.}
\setlength{\tabcolsep}{4pt}
\renewcommand{\arraystretch}{1.05}
\begin{tabular}{llcccccc}
\toprule
\textbf{Dataset} & \textbf{Metric} & \textbf{PReNet} & \textbf{DeepSAD} & \textbf{DevNet} & \textbf{DeepSVDD} & \textbf{BCE} & \textbf{CEDL} \\
\midrule

\multirow{3}{*}{\textbf{Yahoo (25)}} 
 & AUROC & 0.746 & 0.796 & 0.412 & 0.721 & 0.663 & \textbf{0.908} \\
 & AUPR  & 0.255 & 0.314 & 0.016 & 0.203 & 0.529 & \textbf{0.872} \\
 & F1    & 0.324 & 0.340 & 0.046 & 0.265 & 0.586 & \textbf{0.864} \\
\midrule

\multirow{3}{*}{\textbf{KPI (29)}} 
 & AUROC & 0.848 & 0.792 & 0.461 & 0.787 & 0.773 & \textbf{0.893} \\
 & AUPR  & 0.216 & 0.414 & 0.080 & 0.237 & 0.575 & \textbf{0.733} \\
 & F1    & 0.310 & 0.521 & 0.113 & 0.316 & 0.639 & \textbf{0.765} \\
\midrule

\multirow{3}{*}{\textbf{SMAP (5)}} 
 & AUROC & \textbf{0.768} & 0.645 & 0.485 & 0.767 & 0.676 & 0.753 \\
 & AUPR  & 0.202 & 0.177 & 0.080 & 0.257 & 0.310 & \textbf{0.498} \\
 & F1    & 0.255 & 0.237 & 0.148 & 0.450 & 0.393 & \textbf{0.555} \\
\midrule

\multirow{3}{*}{\textbf{MSL (6)}} 
 & AUROC & 0.494 & 0.465 & 0.245 & 0.456 & 0.493 & \textbf{0.640} \\
 & AUPR  & 0.178 & 0.135 & 0.069 & 0.137 & 0.176 & \textbf{0.338} \\
 & F1    & 0.346 & 0.270 & 0.186 & 0.252 & 0.332 & \textbf{0.413} \\
\midrule

\textbf{Rank} 
 & AUROC & 2.000 & 3.500 & 6.000 & 3.750 & 4.250 & \textbf{1.500} \\
 & AUPR  & 3.750 & 4.000 & 6.000 & 4.000 & 2.250 & \textbf{1.000} \\
 & F1    & 3.750 & 3.750 & 6.000 & 4.000 & 2.500 & \textbf{1.000} \\

\bottomrule
\end{tabular}
\label{tab:tsad}
\end{table}
A similar observation can be made from the results in Table \ref{tab:tsad}, where the proposed CEDL method achieves consistently superior performance across all time series benchmarks and metrics, showing its robustness and effectiveness in temporal anomaly detection. Compared with the baseline methods, CEDL yields particularly large gains in AUPR and F1, indicating improved balance between precision and recall under temporal noise and shifting distributions. For instance, on the Yahoo and KPI datasets, CEDL improves AUPR by more than 30\% over the second-best baseline, highlighting its ability to detect subtle and intermittent anomalies in highly dynamic signals.

\subsubsection{Anomaly Detection on Image Data}
\begin{table}[t]
\centering
\caption{Performance comparison on image anomaly detection datasets.
Bold values indicate the best result for each dataset and metric.}
\setlength{\tabcolsep}{4pt}
\renewcommand{\arraystretch}{1.05}
\begin{tabular}{llcccccc}
\toprule
\textbf{Dataset} & \textbf{Metric} & \textbf{PReNet} & \textbf{DeepSAD} & \textbf{DevNet} & \textbf{DeepSVDD} & \textbf{BCE} & \textbf{CEDL} \\
\midrule

\multirow{3}{*}{\textbf{MNIST}} 
 & AUROC & 0.986 & 0.987 & 0.981 & 0.798 & 0.980 & \textbf{0.988} \\
 & AUPR  & 0.918 & 0.915 & 0.915 & 0.429 & 0.893 & \textbf{0.926} \\
 & F1    & 0.838 & 0.837 & \textbf{0.842} & 0.409 & 0.810 & 0.841 \\
\midrule

\multirow{3}{*}{\textbf{F. MNIST}} 
 & AUROC & 0.850 & 0.852 & 0.825 & 0.667 & 0.870 & \textbf{0.895} \\
 & AUPR  & 0.535 & 0.563 & 0.545 & 0.167 & 0.585 & \textbf{0.613} \\
 & F1    & 0.517 & 0.537 & 0.531 & 0.239 & 0.552 & \textbf{0.559} \\
\midrule

\multirow{3}{*}{\textbf{CIFAR10}} 
 & AUROC & 0.821 & 0.803 & 0.816 & 0.562 & 0.822 & \textbf{0.826} \\
 & AUPR  & 0.469 & 0.443 & 0.453 & 0.124 & 0.469 & \textbf{0.474} \\
 & F1    & 0.465 & 0.457 & 0.456 & 0.196 & 0.463 & \textbf{0.475} \\
\midrule

\textbf{Rank} 
 & AUROC & 2.667 & 3.333 & 4.333 & 6.000 & 3.000 & \textbf{1.667} \\
 & AUPR  & 3.000 & 3.333 & 4.667 & 6.000 & 2.667 & \textbf{1.333} \\
 & F1    & 3.000 & 3.333 & 3.667 & 6.000 & 2.667 & \textbf{1.333} \\

\bottomrule
\end{tabular}
\label{tab:image_results}
\end{table}

Similar to our observations on the tabular and time series datasets, the results in Table \ref{tab:image_results} show that CEDL consistently achieves the best overall performance on image anomaly detection, ranking first across all three metrics. On MNIST, CEDL-enabled models yields the highest AUROC and AUPR while maintaining competitive F1 performance. On the more challenging Fashion-MNIST and CIFAR-10 datasets, it outperforms all baselines, highlighting its robustness under higher intra-class variability and real-world visual pattern.

\subsubsection{Overall Anomaly Detection Cross Modalities}
It is worth noting that CEDL demonstrates consistently strong performance across all three data modalities, achieving the highest average ranking in every case. Although baseline approaches can perform well in specific domains, their effectiveness is not consistent across all modalities. For example, DeepSAD and PReNet achieve competitive results on structured tabular datasets but their performance declines on visual or temporal data, where anomaly patterns are more complex and diverse. Similarly, DevNet and BCE classifiers perform reasonably well on image data but tend to suffer from performance degradation on time series benchmarks. This consistent effectiveness demonstrates the robustness of CEDL across diverse data modalities.

\subsection{Ablation Studies}
\subsubsection{Ablation on Anomaly Proportion}
We study the data efficiency of the proposed CEDL loss with its radial logit formulation in comparison to the conventional sigmoid-derived logit (BCE) under different numbers of available anomalies in the training set, as shown in Figure \ref{fig:yahoo_anomaly_proportion_ablation}. We vary the proportion of anomalous labels in the training set (1\%, 5\%, 10\%, 15\%, 20\%) and compare a CEDL head with a BCE head. We report F1, AU-PR and AU-ROC on three Yahoo entities: \textit{real\_15}, \textit{real\_23}, and \textit{real\_24}.

Across entities, CEDL consistently achieves higher F1 and AU-PR than BCE at low anomaly availability (1--5\%), with AU-ROC comparatively similar between heads. As the anomaly proportion increases, both methods generally improve up to $\sim$10--15\%. Behaviour at 20\% is entity-dependent: on \textit{real\_24} we observe a mild dip for CEDL while BCE continues to improve; \textit{real\_15} and \textit{real\_23} tend to plateau or show small additional gains. In particular, we aim to answer the following questions: 

\begin{figure}[htp]
  \centering
  % \subcaptionbox{Yahoo \textit{real\_15}\label{fig:real15}}{%
    \includegraphics[width=1.0\textwidth]{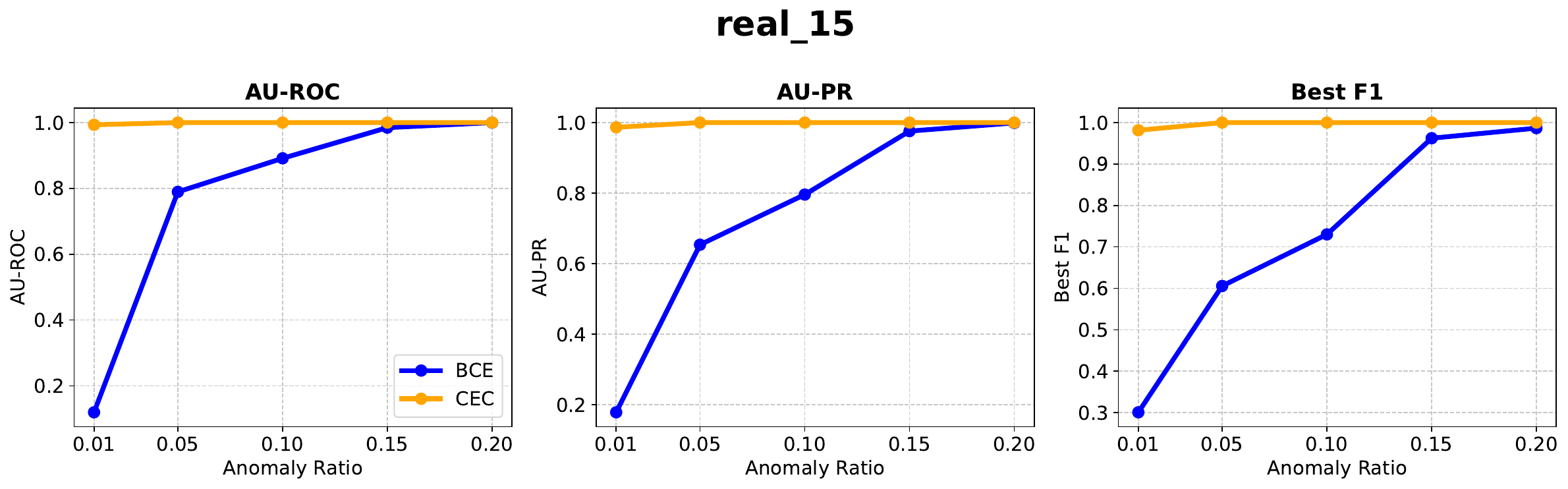}
  % \hfill
  % \subcaptionbox{Yahoo \textit{real\_23}\label{fig:real23}}{%
    \includegraphics[width=1.0\textwidth]{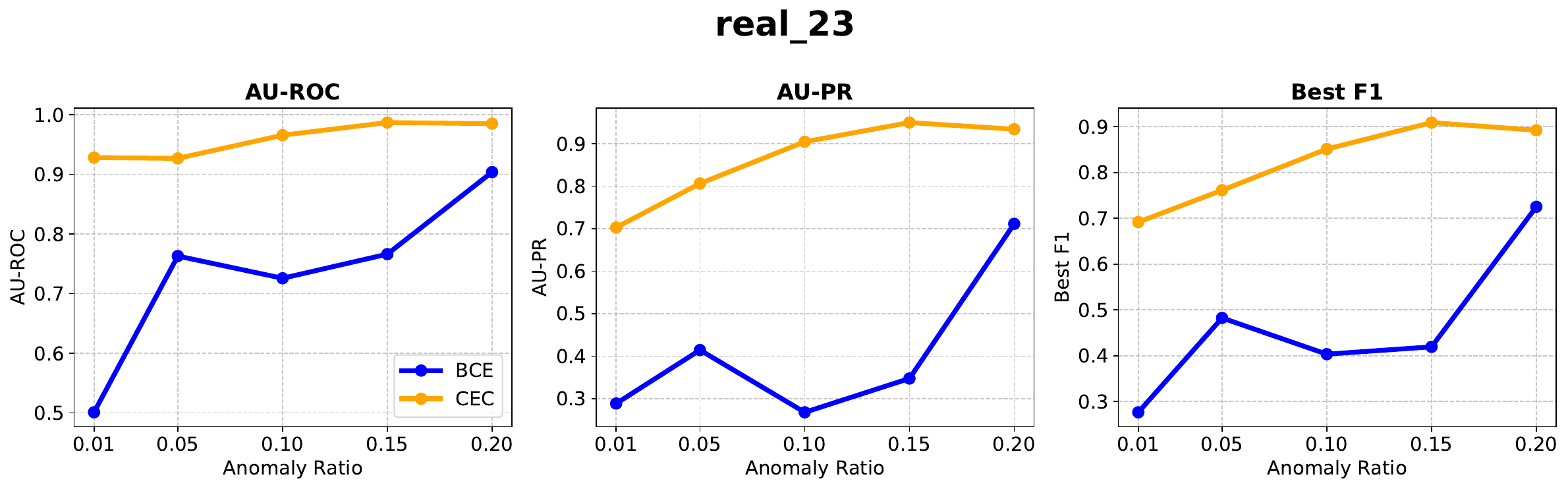}
  % \hfill
  % \subcaptionbox{Yahoo \textit{real\_24}\label{fig:real24}}{%
    \includegraphics[width=1.0\textwidth]{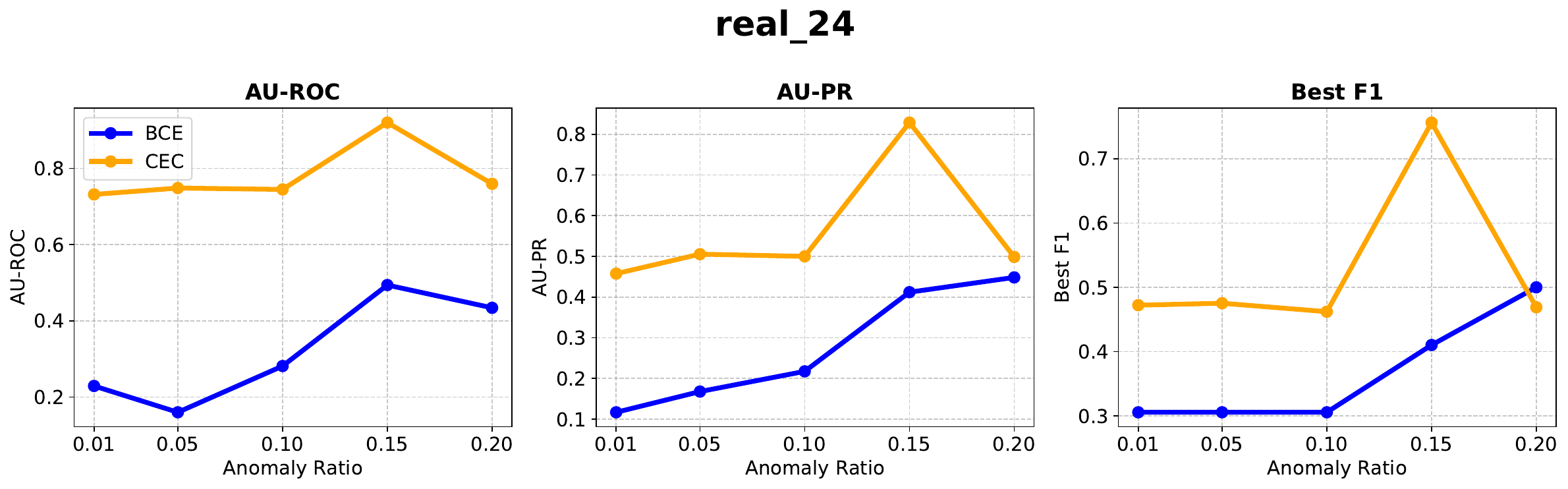}
  \caption{Ablation on training anomaly proportion (1–20\%) for CEDL vs.\ BCE on three Yahoo entities. CEDL leads at low anomaly availability and both models improve toward mid ranges; behaviour at 20\% varies across entities.}
  \label{fig:yahoo_anomaly_proportion_ablation}
\end{figure}

\medskip

\noindent\textbf{Why CEDL excels when anomalies are scarce.} CEDL’s radial logit treats normality as compactness around a learned centre and pushes anomalies outward in all directions. This geometry yields correctly-signed, bounded, and directionally appropriate updates for diverse anomaly modes, which is precisely the regime where few labelled anomalies are available and coverage of sub-modes is sparse---hence markedly better AU-PR/F1 at 1--5\%.

\medskip

\noindent\textbf{Why BCE catches up as anomalies grow.} BCE separates along a single linear projection. With more labelled anomalies, that projection is estimated from richer positive evidence (including minority sub-modes), so BCE’s ranking and thresholded F1/PR improve steadily.

\medskip

\noindent\textbf{Why CEDL can soften at very high proportions.} Once many positives become easy, sigmoid-style saturation reduces their gradient contribution, which can subtly relax the decision rim. In parallel, centre/boundary dynamics may shift if additional anomalies are uneven across sub-modes, nudging some normal regions slightly farther from the centre and introducing a few extra false positives. This explains the small decline for CEDL on \textit{real\_24} at 20\%, even as BCE continues to benefit from the extra positive signal.

\medskip

In realistic deployments, labelled anomalies are rare. The three-entity study shows that under such scarcity CEDL delivers consistently superior AU-PR and F1, while maintaining comparable AU-ROC, making it the more reliable choice for production settings where positive labels are limited and anomaly structure is heterogeneous.

\subsubsection{Ablation on Representation Geometry}
\label{sec:ablation_tsne}

\begin{figure}[t]
  \centering
  \subcaptionbox{CEDL (t-SNE)\label{fig:tsne_cec}}{%
    \includegraphics[width=0.40\textwidth]{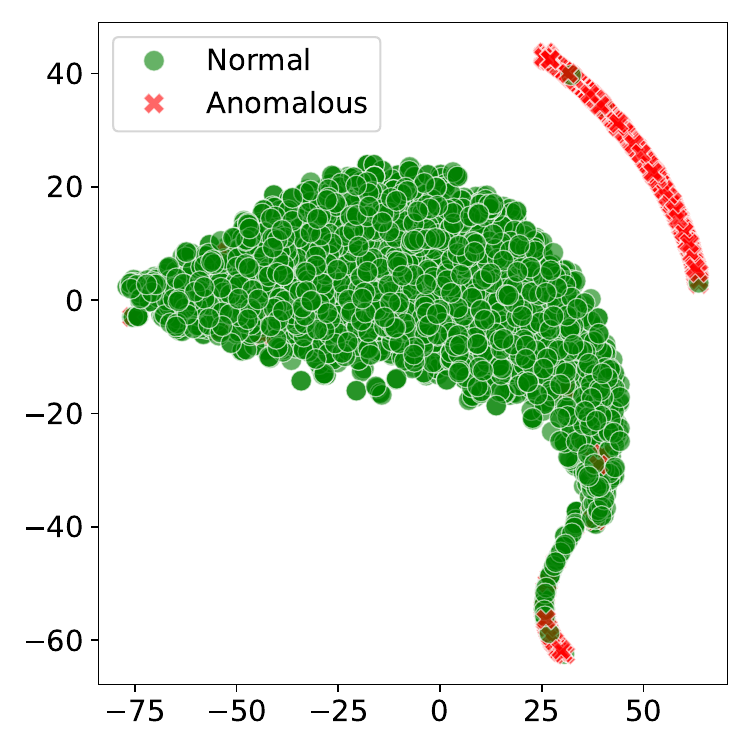}}
  % \hfill
  \subcaptionbox{BCE (t-SNE)\label{fig:tsne_bce}}{%
    \includegraphics[width=0.40\textwidth]{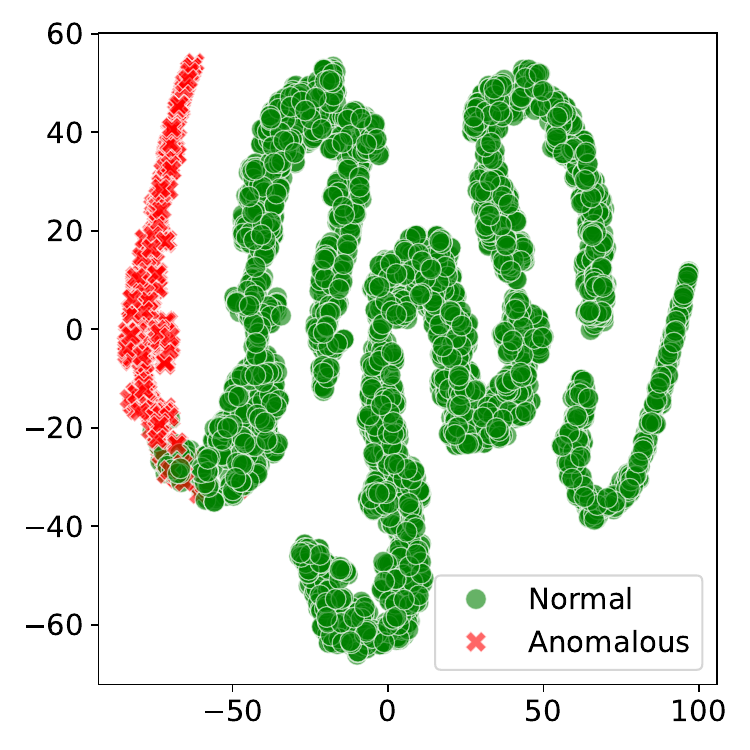}}
  \caption{t-SNE of learned representations of Campaign dataset: CEDL forms a compact normal core with an annulus of anomalies (circular push), whereas BCE exhibits a more elongated, overlapping structure.}
  \label{fig:tsne_cec_bce}
\end{figure}

We project the learned representations of Campaign dataset from the CEDL and BCE heads into 2-D using t-SNE and colour points by class (normal vs.\ anomalous). The visualisations are produced with identical preprocessing and plotting settings for comparability (Figure \ref{fig:tsne_cec_bce}).

The CEDL embedding exhibits a compact normal core surrounded by a pronounced annulus of anomalies—consistent with a \emph{circular outward push} on positives—yielding a clear radial margin between classes. In contrast, the BCE embedding shows a more elongated, partially interleaved layout with broader overlap between normals and anomalies.

CEDL’s radial logit encourages centre-seeking normals and centre-repelling anomalies, producing geometry that is isotropically separable in the representation space. This shape matches the model’s inductive bias (distance-from-centre decisioning) and helps precision–recall by reducing local normal–anomaly mixing. BCE, which optimises a single linear projection, tends to form a stretched band with class overlap along directions not well captured by that projection.

The t-SNE study visually corroborates that CEDL organises features into a compact normal manifold with an anomalous ring (the intended circular push), while BCE yields less radially structured and more overlapping geometry—aligning with CEDL’s stronger PR/F1 in scarce-label regimes.

\section{Conclusion}
In this paper, we introduced Centre-Enhanced Discriminative Learning (CEDL), a novel supervised anomaly detection framework that unifies geometric and discriminative learning within a single end-to-end formulation. By embedding a centre-based radial distance function into the classification objective, CEDL reparameterises the conventional sigmoid-derived logit into a geometry-aware representation of normality. This design enables the model to capture compact normal distributions and produce interpretable, probabilistic anomaly scores without requiring reference data calibration or manual thresholding. Extensive experiments on tabular, time-series, and image anomaly detection benchmarks demonstrate that CEDL achieves consistently competitive and balanced performance across all modalities, validating its effectiveness and broad applicability to diverse real-world application domains.

For future works, we aim to explore several directions to further enhance CEDL’s capabilities. To improve the modelling of the normal class, incorporating multiple centroids could enable the model to better capture the underlying normal class distribution. To broaden its applicability to more diverse anomaly information, extending CEDL to semi-supervised or few-shot learning settings would allow it to perform effectively in scenarios with limited or incomplete labelled data.

\bibliography{refs}

\end{document}